\documentclass[10pt,twocolumn,letterpaper]{article}

\usepackage{cvpr} 
\definecolor{cvprblue}{rgb}{0.21,0.49,0.74}
\usepackage[pagebackref,breaklinks,colorlinks,allcolors=cvprblue]{hyperref}

\def\paperID{29987} 
\def\confName{CVPR}
\def\confYear{2026}

\title{PARL: Position-Aware Relation Learning Network for Document Layout Analysis}

\author{
    Fuyuan Liu\textsuperscript{1}\thanks{Equal contribution.} \quad
    Dianyu Yu\textsuperscript{1,3}\footnotemark[1] \quad
    He Ren\textsuperscript{1} \quad
    Nayu Liu\textsuperscript{4} \quad
    Xiaomian Kang\textsuperscript{2} \quad
    Delai Qiu\textsuperscript{1} \quad
    Fa Zhang\textsuperscript{1} \\ 
    Genpeng Zhen\textsuperscript{1} \quad
    Shengping Liu\textsuperscript{1} \quad
    Jiaen Liang\textsuperscript{1} \quad
    Wei Huang\textsuperscript{1} \quad
    Yining Wang\textsuperscript{1}\thanks{Corresponding author.} \quad
    Junnan Zhu\textsuperscript{2}\footnotemark[2]
    \vspace{0.1em} \\ 
    \textsuperscript{1}Unisound AI Technology Co.Ltd \quad
    \textsuperscript{2}MAIS, Institute of Automation, CAS \\
    \textsuperscript{3}Beihang University \quad
    \textsuperscript{4}School of Computer Science and Technology, Tiangong University \\
    {\tt\small \{junnan.zhu@nlpr.ia.ac.cn, wangyining@unisound.com\}}
}

\begin{document}
\maketitle
\begin{abstract}
Document layout analysis aims to detect and categorize structural elements (e.g., titles, tables, figures) in scanned or digital documents. Popular methods often rely on high-quality Optical Character Recognition (OCR) to merge visual features with extracted text. This dependency introduces two major drawbacks: propagation of text recognition errors and substantial computational overhead, limiting the robustness and practical applicability of multimodal approaches. In contrast to the prevailing multimodal trend, we argue that effective layout analysis depends not on text-visual fusion, but on a deep understanding of documents' intrinsic visual structure. To this end, we propose PARL (Position-Aware Relation Learning Network), a novel OCR-free, vision-only framework that models layout through positional sensitivity and relational structure. Specifically, we first introduce a Bidirectional Spatial Position-Guided Deformable Attention module to embed explicit positional dependencies among layout elements directly into visual features. Second, we design a Graph Refinement Classifier (GRC) to refine predictions by modeling contextual relationships through a dynamically constructed layout graph. Extensive experiments show PARL achieves state-of-the-art results. It establishes a new benchmark for vision-only methods on DocLayNet and, notably, surpasses even strong multimodal models on M6Doc. Crucially, PARL (65M) is highly efficient, using roughly four times fewer parameters than large multimodal models (256M), demonstrating that sophisticated visual structure modeling can be both more efficient and robust than multimodal fusion.

\end{abstract}    
\section{Introduction}
\label{sec:intro}

Document layout analysis aims to detect and categorize semantic regions within document images, such as titles, paragraphs, tables, and figures~\cite{binmakhashen2019document,DBLP:journals/algorithms/FischerHP23}. As a fundamental component in document intelligence systems, its performance directly impacts downstream applications like information extraction and semantic retrieval~\cite{DBLP:conf/icdar/SciusBertrandFVCF24}.

\begin{figure}
\centering
\includegraphics[width=80mm,page=1]{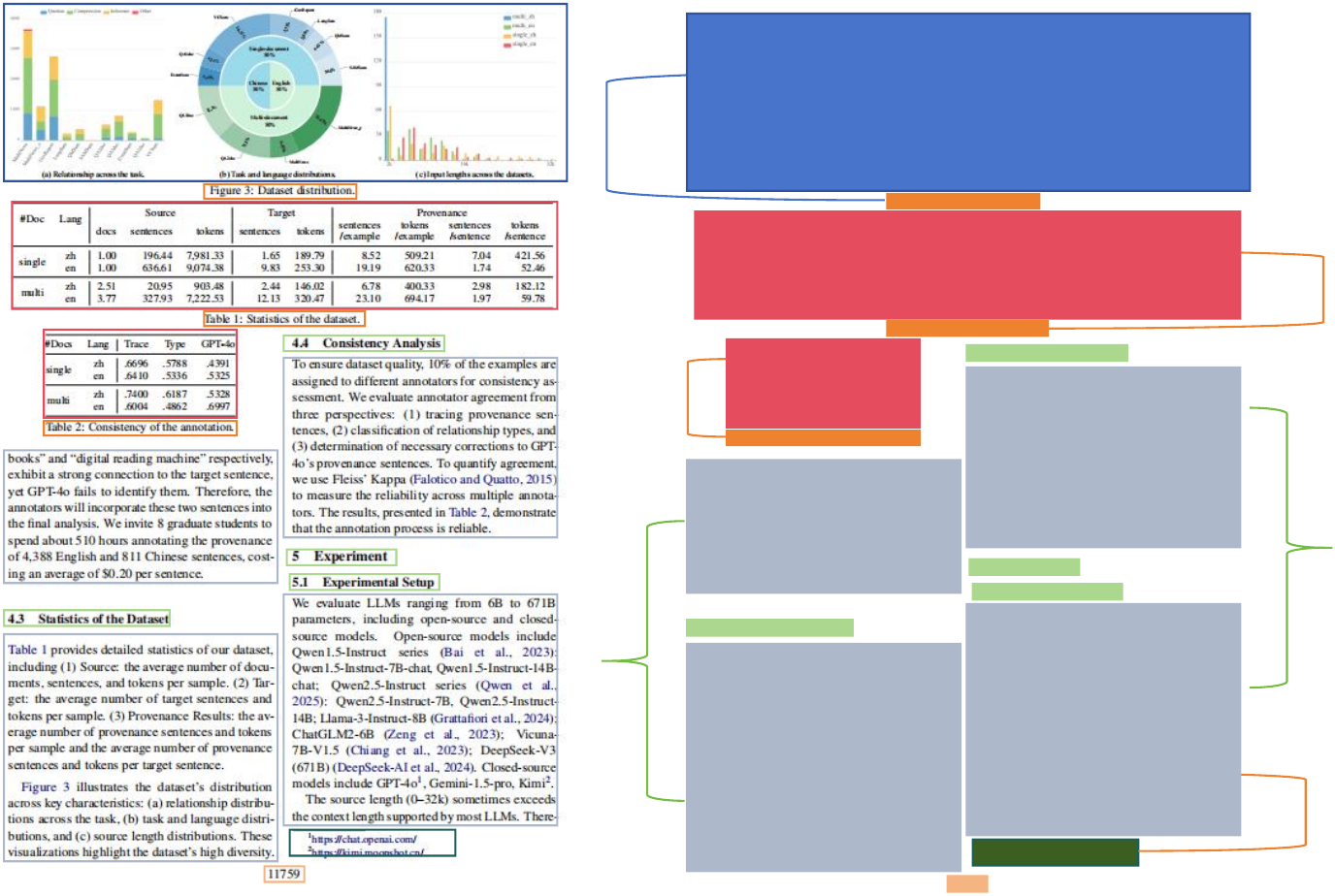}
\caption{As illustrated, this figure shows the ``visual grammar'' of document layout: elements do not exist in isolation but are closely correlated with their context and their own spatial position.}
\vspace{-0.5cm}
\end{figure}

With the development of OCR, multimodal approaches have been continuously studied, mainly focusing on the effective integration of OCR-based text and visual features~\cite{zhang2024document,shehzadi2024hybrid}. Despite being able to utilize rich textual information, these approaches face inevitable computational overhead and error propagation issues caused by OCR systems, which limit their application in resource-constrained real-time scenarios. Therefore, OCR-free pure-visual approaches have recently received increasing attention~\cite{tewes2025sfdla}. This type of method models the task as a multi-objective detection task in CV and applies general-purpose detectors to identify regions in isolation. However, unlike natural scenes with stochastic object arrangements, document images adhere to strong structural regularities, an implicit ``visual grammar,'' which are ignored by current approaches.

This ``visual grammar'' is the crux of the problem. Crucially, this grammar is spatial and structural, not linguistic, creating strong priors that link an element's category to its spatial context, a structure discernible even without reading the text. We argue this grammar is defined by two key components: (1) Positional and Spatial Priors, where an element's absolute and relative location strongly suggests its identity (e.g., a page number is almost always found at the absolute edge of the layout, such as the footer or header.), and (2) Inter-Element Context, where an element's class is confirmed by its relationship with its neighbors (e.g., a figure caption or table caption is typically positioned immediately adjacent to the visual it describes). Existing vision-based methods fail to model both of these components effectively, creating a distinct performance gap.

Grounded in the insight that this ``visual grammar'' is critical for robust layout analysis, we introduce \textbf{PARL (Position-Aware Relation Learning Network)}, a specialized pure-vision framework that systematically models document structure by integrating relational reasoning at two complementary levels: the feature-level and the decision-level. Our approach is grounded in the hypothesis that this dual-level modeling of spatial relationships can effectively compensate for the absence of textual information. PARL achieves this through two complementary mechanisms: (1) \textbf{Feature-Level Relational Modeling}: We propose a Bidirectional Spatial Position-Guided Deformable Attention (BSP-DA) module. This module operates within the decoder, leveraging the bidirectional spatial relationships between all object queries to dynamically compute a global relational adjustment for its sampling offsets. This embeds structural priors directly into the feature extraction process, making the resulting visual representations inherently grammar-aware. (2) \textbf{Decision-Level Relational Reasoning}: We introduce a Graph Refinement Classifier (GRC) that acts as a refinement step. Rather than a standard, isolated classification head, the GRC constructs a global topological graph from the decoder's output embeddings and predicted bounding boxes. It then executes a graph attention network to perform explicit relational reasoning at the decision-level, reasoning over the \textit{entire} layout to ensure the final predictions are globally and structurally coherent.

Extensive experiments on public benchmarks (M6Doc, DocLayNet, and D4LA) validate PARL's strong performance. It achieves state-of-the-art (SOTA) on M6Doc and D4LA, and establishes a new SOTA among vision-only methods on DocLayNet. Notably, our parameter-efficient model surpasses several computationally intensive multimodal methods, validating our approach as a viable path toward high-performance, lightweight document intelligence.

Our main contributions are summarized as follows:
\begin{itemize}
 \item We propose PARL, a pure-vision framework built upon the ``visual grammar'' intuition. It is designed to systematically capture the structural and spatial conventions of documents by integrating relational reasoning at both the feature-level and the decision-level.
 \item To implement this approach, we introduce two novel architectural components: (1) the Bidirectional Spatial Position-Guided Deformable Attention (BSP-DA) to embed relational priors at the feature-level, and (2) the Graph Refinement Classifier (GRC), which constructs a dynamic feature-spatial graph and performs decision-level reasoning through an adaptive fusion of baseline and graph-refined predictions.
 \item We demonstrate SOTA performance on M6Doc and D4LA, and a new vision-only SOTA on DocLayNet. PARL outperforms numerous multimodal methods, offering a superior trade-off between accuracy and efficiency.
\end{itemize}
\section{Related Work}
\label{sec:formatting}
\subsection{Document Layout Analysis}
Automated analysis of document layouts is a long-standing research problem. With the advent of deep learning, approaches have largely split into two categories: vision-based and multimodal methods.

\textbf{Vision-based Methods.} This line of research treats document layout analysis as a purely visual task, often reformulating it as an object detection or semantic segmentation problem. Standard object detectors, such as Faster R-CNN and Mask R-CNN~\cite{oliveira2017fast, xu2021page}, supported by toolkits like LayoutParser~\cite{shen2021layoutparser}, and specialized architectures like Doclayout-YOLO~\cite{zhao2024doclayout}, have been successfully adapted for this purpose. More recently, Transformer-based models, such as DLAFormer~\cite{wang2024dlaformer}, attempt to unify subtasks into a relation prediction problem. However, these vision-based approaches predominantly model layout elements in isolation, lacking an explicit mechanism to capture complex inter-object relationships and the global document structure—a limitation our work directly addresses.

\textbf{Multimodal Methods.}A paradigm shift was catalyzed by pre-trained language models, which fuse visual information with OCR-extracted text for a deeper semantic understanding. The seminal LayoutLM series~\cite{xu2020layoutlm, xu2020layoutlmv2,huang2022layoutlmv3} pioneered this by augmenting a BERT-like architecture with 2D position embeddings and visual features. Subsequent works have built upon this transformer-based foundation, introducing innovations such as vision-centric pre-training~\cite{li2022dit}, grid-based feature encoding, plug-in fusion modules~\cite{zhang2024m2doc}, and language-independent frameworks~\cite{wang2022lilt}. A distinct line of research focuses on explicitly modeling relationships using graph structures; Doc-GCN~\cite{DBLP:conf/coling/LuoDLPH22}, for instance, constructs a heterogeneous graph to integrate features and learn contextualized representations. More recently, methods like LayoutLLM~\cite{luo2024layoutllm} adapt general-purpose Large Language Models via layout-aware instruction tuning, demonstrating strong zero-shot capabilities. While these models have set new state-of-the-art benchmarks, they often depend on an OCR engine, making them susceptible to cascading errors. Furthermore, the very architecture required to fuse visual features with this textual information typically involves large, transformer-based language models, which introduces a substantial computational overhead. Our work is motivated by achieving their level of contextual understanding without incurring the substantial computational cost.

\subsection{Relational Reasoning in Object Detection}

\begin{figure*}[htbp]
\centering
\includegraphics[scale=0.46]{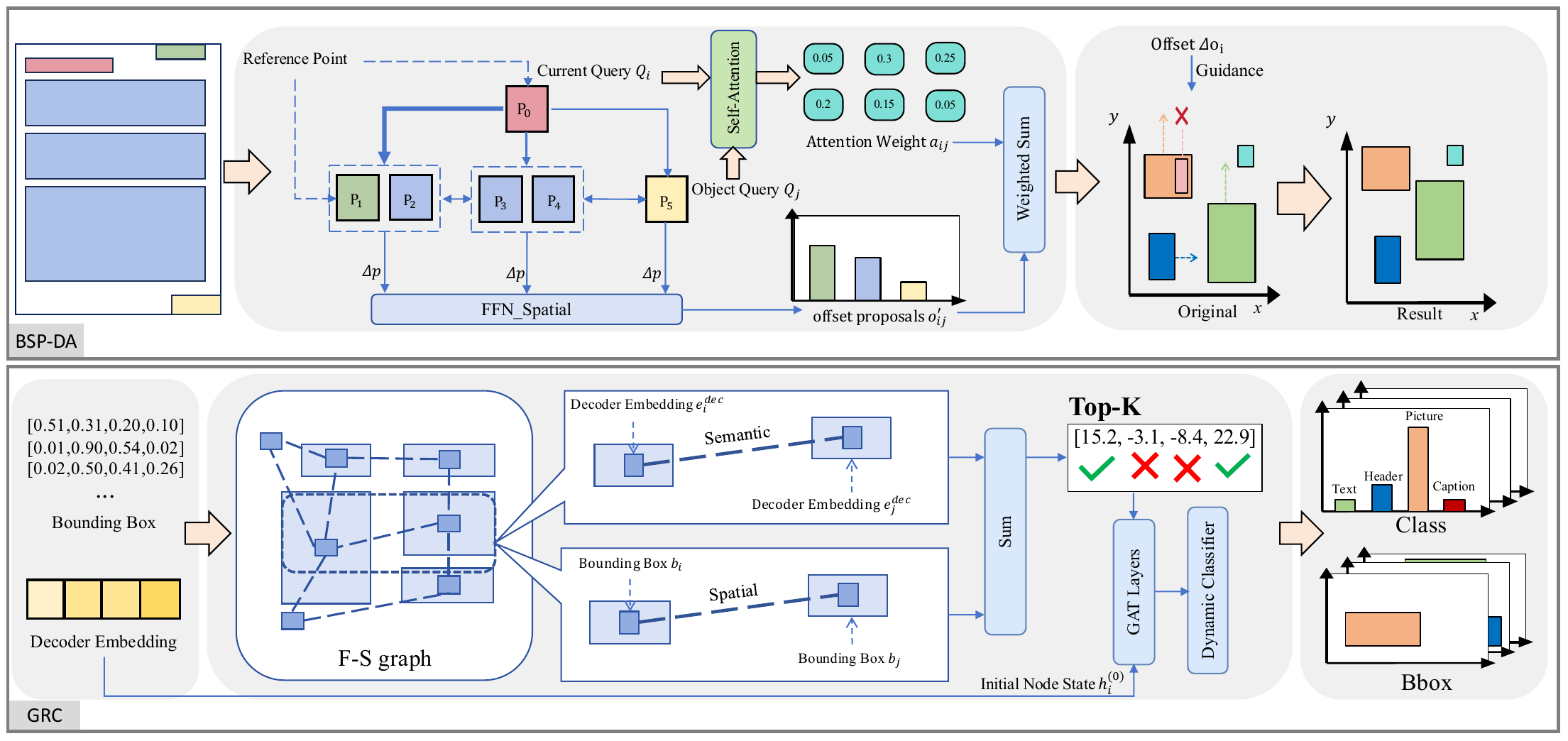}
\caption{Overview of our proposed method. The BSP-DA module injects spatial priors into the attention mechanism. It learns from global spatial relationships (via FFN Spatial) and semantic context (via Self-Attention) to compute an ``Offset $\Delta \boldsymbol{o}_i$ Guidance,'' which corrects the ``Original'' sampling points into a more precise "Result." The GRC module refines the final classification by modeling contextual relationships. It constructs an F-S graph based on Semantic and Spatial cues, and then uses this graph context (via GAT Layers and a Dynamic Classifier) to optimize the final class predictions.}
\label{figure}
\vspace{-0.5cm}
\end{figure*}

Conventional object detectors are designed to localize and classify individual instances. They largely ignore the relationships between these instances. Recognizing this, a significant body of research has focused on incorporating relational reasoning into detection frameworks. This concept is central to holistic scene understanding. Early efforts included methods like Relation Networks~\cite{santoro2017simple}. These employ a generic function to reason about pairs of object proposals.

Drawing inspiration from these explicit relational priors, our work agrees they are key. However, we specialize this concept for the unique, highly structured domain of document layout analysis. Furthermore, our implementation is fundamentally different from general-purpose detectors like Relation DETR~\cite{hou2024relation}. Instead of modifying the self-attention logits, PARL introduces a dual-level relational framework: \textbf{(1) The BSP-DA module} dynamically modulates the sampling offsets of attention based on a global layout understanding. \textbf{(2) The GRC module} performs explicit relational reasoning at the decision level to refine classifications. This dual mechanism is specifically designed to disambiguate visually similar document elements (like titles vs. paragraphs) by reasoning about their structure—a challenge not addressed by general-purpose detectors.

\section{Proposed Method}
\label{sec:method}
We introduce PARL (Position-Aware Relation Learning network), a novel framework for efficient and accurate document layout analysis. Our method is built upon the DFINE multi-object detection framework (Sec. 3.1) and enhances it by explicitly modeling the visual grammar and relational structure inherent in document layouts. This is achieved using two synergistic innovations. First, we propose the Bidirectional Spatial Position-Guided Deformable Attention (BSP-DA) module (Sec. 3.2). This module provides more precise guidance for the model's attention range by referencing the global relative spatial relationships between elements. Second, we design the Graph Refinement Classifier (GRC) (Sec. 3.3). The GRC dynamically constructs an adjacency graph using both visual position and semantic information, thereby providing crucial contextual information to improve the final class prediction for each element.

\subsection{Preliminaries}

We formulate document layout analysis as a specialized multi-object detection task, one that is governed by strong structural and spatial constraints. In this domain, the DETR~\cite{carion2020end} paradigm is a foundational approach, modeling the task as an end-to-end set prediction problem. It utilizes a set of $N$ learnable object queries, $\mathcal{Q} = \{\boldsymbol{q}_i\}_{i=1}^N$. After an image is processed by a feature extractor, each query $\boldsymbol{q}_i$ probes the resulting visual tokens using a Transformer decoder. The output embedding is then fed to parallel heads to predict the class and bounding box coordinates.

Building on this, Deformable DETR~\cite{ZhuSLLWD21} introduced Deformable Attention, which replaces the standard Transformer attention. Our work is built upon DFINE~\cite{peng2024d}, a strong successor to Deformable DETR. The core innovation of Deformable Attention is its ability to focus on a small, dynamic set of $K$ key sampling points around a reference point $\boldsymbol{p}_i$, rather than attending to all tokens. This allows the model to learn \textit{where} to look. The output feature is a weighted sum of sampled features from the feature map $\boldsymbol{x}$:
\begin{equation}
\label{eq:deform_attn}
\small
\text{DeformAttn}(\boldsymbol{q}_i, \boldsymbol{p}_i, \boldsymbol{x}) = \sum_{k=1}^{K} A_{ik} \cdot \boldsymbol{x}(\boldsymbol{p}_i + \Delta \boldsymbol{p}_{ik})
\end{equation}
where $A_{ik}$ are the attention weights and $\Delta \boldsymbol{p}_{ik}$ are the learned offsets for the $k$-th sampling point.

Standardly, the set of $K$ offsets (which we denote collectively as $\boldsymbol{o}_i^{\text{base}}$ to align with our notation in Sec. 3.2) and the $K$ attention weights (denoted as $\boldsymbol{A}_i$) are generated solely from the query $\boldsymbol{q}_i$ via linear projections:
\begin{equation}
\label{eq:bsp_base_o_a}
\small
\begin{gathered}
\boldsymbol{o}_i^{\text{base}} = f_{\text{base}}(\boldsymbol{q}_i) \\
\boldsymbol{A}_i = \text{Softmax}(f_{\text{attn\_weights}}(\boldsymbol{q}_i))
\end{gathered}
\end{equation}

However, a key limitation of these frameworks is that each query $\boldsymbol{q}_i$ is predicted independently, both in the attention calculation (Eq.~\ref{eq:bsp_base_o_a}) and in the final classification head, failing to capture the "visual grammar" and relational priors where an element's identity depends on its neighbors.

\subsection{Bidirectional Spatial Position-Guided Deformable Attention}
Document layouts are governed by strong spatial priors. Standard deformable attention, however, learns offsets for each query individually, failing to leverage this global structure. This limitation is fundamental, as the sampling offsets $\boldsymbol{o}_i$ are generated solely from the query $\boldsymbol{q}_i$ itself, rendering the decision process isolated. The model lacks a built-in mechanism to enforce correlated sampling based on the global relative spatial arrangement between queries.

To address this, we designed the Bidirectional Spatial Position-Guided Deformable Attention (BSP-DA) module,which dynamically fuses the final sampling offset $\boldsymbol{o}_i^{\text{final}}$ should be dynamically fused from two components: (1) a base offset $\boldsymbol{o}_i^{\text{base}}$ which is content-based and query-specific, identical to the standard approach; and (2) our novel relational offset adjustment $\Delta \boldsymbol{o}_i$ which encodes global query-to-query spatial relationships.

For each query $\boldsymbol{q}_i$, its final structurally-aware sampling offset $\boldsymbol{o}_i^{\text{final}}$ is a weighted sum of the base offset and the relational adjustment. We introduce a scalar gating factor $\lambda_i \in [0,1]$, generated from the query $\boldsymbol{q}_i$ itself, to adaptively modulate the strength of the global structural prior.

The complete fusion process is defined as:
\begin{equation}
\label{eq:bsp_final_o}
\small
\boldsymbol{o}_i^{\text{final}} = \boldsymbol{o}_i^{\text{base}} + \lambda_i \cdot \Delta \boldsymbol{o}_i
\end{equation}
Here, the gating factor $\lambda_i$ is derived from the query $\boldsymbol{q}_i$ via a linear layer followed by a Sigmoid activation, computed as $\lambda_i = \sigma(\boldsymbol{W}_{\lambda} \boldsymbol{q}_i + \boldsymbol{b}_{\lambda})$.

This design allows the model to autonomously decide, based on the query's context, how much to rely on the content-driven base offset (when $\lambda_i \to 0$) versus the structure-driven relational adjustment (when $\lambda_i \to 1$).

\textbf{Global Relational Offset Adjustment.}
The relational offset adjustment $\Delta \boldsymbol{o}_i$ is designed to correct the base offset by leveraging the spatial relationships between all queries. We define $\Delta \boldsymbol{o}_i$ as the weighted average of all "offset proposals" $\boldsymbol{o}'_{ij}$ from other queries $j$ to query $i$:
\begin{equation}
\label{eq:bsp_delta_o}
\small
\Delta \boldsymbol{o}_i = \sum_{j=1}^{N} \alpha_{ij} \boldsymbol{o}'_{ij}
\end{equation}
The offset proposal $\boldsymbol{o}'_{ij}$ aims to learn a structured spatial mapping: if two queries have a specific spatial relationship (e.g., $i$ is "below and to the right of" $j$), this term learns how the offset of $i$ should be adjusted.

To learn this mapping, we first compute the raw displacement $\Delta \boldsymbol{p}_{ij}$. This displacement is the vector difference $\boldsymbol{p}_i - \boldsymbol{p}_j$, where $\boldsymbol{p}_i$ and $\boldsymbol{p}_j$ are the normalized 2D reference points (i.e., predicted box centers associated with queries $\boldsymbol{q}_i$ and $\boldsymbol{q}_j$, respectively, as introduced in Sec. 3.1). This raw displacement is then transformed into a concrete offset proposal $\boldsymbol{o}'_{ij}$ using a Feed-Forward Network (FFN), implemented as a multi-layer perceptron:
\begin{equation}
\label{eq:bsp_offset_proposal}
\small
\boldsymbol{o}'_{ij} = \text{FFN}_{\text{spatial}}(\Delta \boldsymbol{p}_{ij})
\end{equation}
Through this process, $\boldsymbol{o}'_{ij}$ effectively learns a complex function mapping a "relative spatial position" to a "suggested offset adjustment."

The weight $\alpha_{ij}$ quantifies how much query $i$ should "listen" to the proposal $\boldsymbol{o}'_{ij}$ from query $j$ when aggregating its relational offset. We compute these weights via a query-to-query self-attention mechanism. Specifically, we use standard scaled dot-product attention to compute an alignment score $e_{ij}$ between queries $i$ and $j$, which is then normalized via Softmax:
\begin{equation}
\label{eq:bsp_attn_scores}
\small
\begin{gathered}
e_{ij} = \frac{(\boldsymbol{q}_i \boldsymbol{W}_Q) (\boldsymbol{q}_j \boldsymbol{W}_K)^T}{\sqrt{d_k}} \\
\alpha_{ij} = \frac{\exp(e_{ij})}{\sum_{l=1}^{N} \exp(e_{il})}
\end{gathered}
\end{equation}
This allows the computation of $\Delta \boldsymbol{o}_i$ to dynamically focus on the queries most relevant to $\boldsymbol{q}_i$.

\subsection{Graph Refinement Classifier}
Standard object detectors classify each object in isolation, overlooking the critical insight that an element's class is tightly coupled with its neighbors. To overcome this limitation, we designed the Graph Refinement Classifier (GRC), which explicitly models these internal, implicit relationships. The GRC operates by constructing a dynamic graph from the set of predicted elements, employing a message passing mechanism to refine each element's features with rich contextual information from its neighbors.

\textbf{Feature-Spatial Graph Construction.} We posit that graph construction based on rigid, purely spatial metrics (e.g., graphs built using only spatial proximity) is suboptimal. Such methods are agnostic to node content and may erroneously connect semantically unrelated elements (e.g., a page number to a table cell) merely due to proximity.

To address this, our GRC dynamically constructs the graph $\mathcal{G} = (\mathcal{V}, \mathcal{E})$ using an adaptive adjacency mechanism that jointly models semantic affinity and spatial proximity.

For each potential edge from node $i$ to node $j$, we compute a joint relational score $s_{ij}$ by fusing two components:

1.Semantic Affinity ($s_{ij}^{\text{feat}}$): We measure content similarity using the decoder features. A learned, query-key dot-product attention is used to compute the semantic relevance:
\begin{equation}\label{eq:grc_feat_score}
\small
s_{ij}^{\text{feat}} = \frac{(\boldsymbol{e}_i^{\text{dec}} \boldsymbol{W}_Q)^T (\boldsymbol{e}_j^{\text{dec}} \boldsymbol{W}_K)}{\sqrt{d_k}}
\end{equation}
where $\boldsymbol{e}_i^{\text{dec}}$ and $\boldsymbol{e}_j^{\text{dec}}$ are the decoder embeddings, and $(\boldsymbol{W}_Q, \boldsymbol{W}_K)$ are learnable projection matrices.

2.Spatial Proximity ($s_{ij}^{\text{spatial}}$): We explicitly model the relative geometry between the two elements. The bounding boxes $(\boldsymbol{b}_i, \boldsymbol{b}_j)$ are passed through a relative positional encoding function $f_{\text{spatial}}$ (e.g., an MLP operating on the normalized coordinate differences) to produce a high-dimensional spatial embedding, which is then projected to a scalar score:
\begin{equation}\label{eq:grc_spatial_score}
\small
s_{ij}^{\text{spatial}} = \boldsymbol{w}_{\text{spatial}}^T f_{\text{spatial}}(\boldsymbol{b}_i, \boldsymbol{b}_j)
\end{equation}
where $f_{\text{spatial}}$ is the encoding MLP and $\boldsymbol{w}_{\text{spatial}}$ is a learnable weight vector.

The final adjacency score $s_{ij}$ simply sums these two factors:
\begin{equation}\label{eq:grc_adj_score}
\small
s_{ij} = s_{ij}^{\text{feat}} + s_{ij}^{\text{spatial}}
\end{equation}
A directed edge $(i, j)$ is added to the edge set $\mathcal{E}$ if node $j$ is among the Top-K nodes ranked by the score $s_{ij}$ (where $i$ is the source). This "feature-spatial KNN" approach ensures that the resulting neighborhood $\mathcal{N}_i$ for each node is both spatially close and semantically relevant.

\textbf{Node Feature Enhancement.} To ensure the GNN is geometrically aware, we enrich the initial node features. The decoder embedding $\boldsymbol{e}_i^{\text{dec}}$ (also used in semantic affinity scoring) is concatenated with a learned representation of its bounding box $\boldsymbol{b}_i$ (via an MLP $f_{\text{box}}$). The result is projected by $f_{\text{proj}}$ to create the initial node state $\boldsymbol{h}_i^{(0)}$:
\begin{equation}
\label{eq:grc_h0}
\small
\boldsymbol{h}_i^{(0)} = f_{\text{proj}} \left( \left[ \boldsymbol{e}_i^{\text{dec}} \Vert f_{\text{box}}(\boldsymbol{b}_i) \right] \right)
\end{equation}
where $[\cdot \Vert \cdot]$ denotes concatenation.

\textbf{Graph Attention Propagation.} We use a multi-layer GAT to refine node features. Each node aggregates information from its neighbors $\mathcal{N}_i$, weighted by attention coefficients $\alpha_{ij}^{(l)}$:
\begin{equation}
\label{eq:grc_gat}
\small
\boldsymbol{h}_i^{(l+1)} = \sigma\left(\sum_{j \in \mathcal{N}_i \cup \{i\}} \alpha_{ij}^{(l)} \boldsymbol{W}^{(l)} \boldsymbol{h}_j^{(l)}\right)
\end{equation}
where $\boldsymbol{W}^{(l)}$ is a learnable weight matrix and $\sigma$ is a non-linear activation.

\textbf{Final Classification.} The final classification score is not derived from the GAT alone. Instead, it is a dynamic interpolation between a baseline classifier ($\mathrm{Score}_{\text{base}}$), which operates on the original decoder features $\boldsymbol{e}_i^{\text{dec}}$, and a GAT-based classifier ($\mathrm{Score}_{\text{GAT}}$), which operates on the refined features $\boldsymbol{h}_i^{(L)}$ from the final GAT layer:
\begin{equation}
\label{eq:grc_score}
\small
\mathrm{Score}_{\text{final}} = \alpha \cdot \text{FFN}_{\text{base}}(\boldsymbol{e}_i^{\text{dec}}) + (1-\alpha) \cdot \text{FFN}_{\text{GAT}}(\boldsymbol{h}_i^{(L)})
\end{equation}
Where $\alpha$ is a learnable scalar that adaptively balances the contribution of the two classifiers.

\subsection{Loss Function}
We use a set-based loss with bipartite matching. The loss combines $\mathcal{L}_{\text{box}}$ (L1 loss) and $\mathcal{L}_{\text{giou}}$ (Generalized IoU loss) for regression. For classification, we use Varifocal Loss (VFL)~\cite{zhang2021varifocalnet} to address class imbalance. VFL weights positive samples by their predicted IoU, compelling the model to prioritize high-quality localizations. The overall loss is a weighted sum:
\begin{equation}
\label{eq_loss}
\small
\mathcal{L} = \lambda_{\text{vfl}}\mathcal{L}_{\text{vfl}} + \lambda_{\text{box}}\mathcal{L}_{\text{box}} + \lambda_{\text{giou}}\mathcal{L}_{\text{giou}}
\end{equation}
where $\lambda_{\text{vfl}}$, $\lambda_{\text{box}}$, and $\lambda_{\text{giou}}$ are loss coefficients.
\section{Experiments}
\subsection{Datasets}
Our experiments are conducted on three widely used public datasets for document layout analysis:
\begin{itemize}
    \item \textbf{D4LA}~\cite{da2023vgt} is a fine-grained document layout dataset comprising 27 distinct categories. It is designed to evaluate a model's capability for detailed and complex layout analysis in diverse scenarios.
    \item \textbf{DocLayNet}~\cite{pfitzmann2022doclaynet} is a comprehensive large-scale benchmark that consists of over 80,000 pages that span 11 distinct categories, representing a significant challenge for fine-grained layout recognition due to its structural diversity and hierarchical complexity.
    \item \textbf{M6Doc}~\cite{cheng2023m6doc} is a multi-format, multi-lingual, and multi-domain document layout dataset. We utilize it to assess the generalization and robustness of our model across varied document styles.
\end{itemize}

\subsection{Evaluation Metrics and Implementation Details}
\noindent\textbf{Implementation Details.} PARL is built in PyTorch, utilizing an HGNetv2~\cite{ma2019paddlepaddle} backbone pretrained on ImageNet~\cite{deng2009imagenet}. All models are trained on four NVIDIA 4090 GPUs. We use the AdamW~\cite{loshchilov2017decoupled} optimizer with a base learning rate of $2.5 \times 10^{-4}$. The global weight decay is set to $1.25 \times 10^{-4}$, from which normalization layers are exempted. Since the original publications for the competing methods often do not report efficiency metrics (e.g., Latency and GFLOPs), we re-evaluated all methods within our unified codebase to ensure a rigorous and fair comparison. All reported efficiency metrics (Params, Latency, and GFLOPs) were benchmarked on a single NVIDIA 4090 GPU, with latency measured using a batch size of 1.

\subsection{Results and Discussion}
In this section, we present a comprehensive evaluation of our proposed model, PARL. Our experiments are designed to rigorously assess its performance against our baseline architecture, DFINE-X, and other state-of-the-art (SOTA) models. We conduct comparisons on the \textbf{M6Doc, DocLayNet, and D4LA} benchmarks, with particular attention to model efficiency in terms of parameter count.

\noindent\textbf{Performance on M6Doc.}
As detailed in Table~\ref{tab:m6doc_results}, PARL sets a new state-of-the-art on the challenging M6Doc dataset. Our model achieves an mAP of \textbf{72.9\%}, surpassing strong models based on pure vision, including our baseline DFINE-X (70.3\%) and DINO (68.0\%), with significant margins of 2.6 and 4.9 points, respectively. In particular, PARL outperforms sophisticated models that leverage \textbf{ multimodal fusion}, such as DINO (m2doc) (69.9\%), highlighting the efficacy of explicitly modeling intrapage relational structures. PARL leads in all primary metrics, reaching the highest $AP_{50}$ (86.8\%) and AR (83.7\%). Importantly, this superior performance is achieved with a highly efficient parameter count of only 65M, comparable to DINO (67M) but substantially smaller than larger architectures like VSR (306M) and Cascade (m2doc) (285M). These results collectively validate that our approach of integrating explicit positional and relational priors offers a highly effective and efficient solution for complex document layout analysis.

\begin{table}[t]
\small
\centering
\setlength{\tabcolsep}{0.5mm} 
\begin{tabular}{l|ccc|ccc}
\toprule
\textbf{Method} & \textbf{Params} & \textbf{Latency} & \textbf{GFLOPs} & \textbf{AP50} & \textbf{AR} & \textbf{mAP} \\
\midrule
\multicolumn{7}{c}{\textit{Multimodal Methods}} \\
\midrule
VSR & 306M & - & - & 76.2 & 66.4 & 59.9 \\
Cascade (m2doc) & 285M & 95.96 & 907 & 78.0 & 67.9 & 61.8 \\
DINO (m2doc) & 256M & 87.25 & 770 & \underline{86.7} & 82.5 & 69.9 \\
DoPTA & - & - & - & 85.8 & - & 69.5 \\
\midrule
\multicolumn{7}{c}{\textit{Vision-only Methods}} \\
\midrule
SOLOv2 & 69M & 33.24 & 223 & 67.5 & 61.5 & 46.8 \\
Faster R-CNN & 60M & 26.28 &181 & 67.8 & 57.2 & 49.0 \\
Mask R-CNN & 63M & 43.08 & 290 & 58.4 & 50.8 & 40.1 \\
Cascade & 96M & 52.93 & 286 & 70.5 & 62.1 & 54.4 \\
HTC & 99M & 73.68 & 597 & 74.3 & 68.1 & 58.2 \\
SCNet & 114M & 84.63 & 681 & 73.5 & 67.3 & 56.1 \\
Deformable DETR & 59M & 31.09 & 173 & 76.8 & 75.2 & 57.2 \\
QueryInst & 192M & 49.78 & 139 & 67.1 & 71.0 & 51.0 \\
TransDLANet & - & - & - & 82.7 & 74.9 & 64.5 \\
DINO & 67M & 34.89 & 279 & 84.6 & 82.9 & 68.0 \\
DocLayout-YOLO & 26M & 13.05 & 68 & 82.1 & 78.2 & 66.5 \\
DFINE-X & 62M & 22.47 & 202 & 84.5 & \underline{83.0} & \underline{70.3} \\
\hline
\textbf{PARL (ours)} & 65M & 28.65 & 217 & \textbf{86.8} & \textbf{83.7} & \textbf{72.9} \\
\bottomrule
\end{tabular}
\caption{Performance comparisons on M6Doc. The best and second best results are shown in \textbf{bold} and \underline{underline}. }
\label{tab:m6doc_results}
\vspace{-0.5cm}
\end{table}

\noindent\textbf{Performance on DocLayNet.}
To further assess its robustness, we evaluate PARL on the diverse and complex layouts of the DocLayNet dataset. The results, summarized in Table~\ref{tab:doclaynet_results}, corroborate our findings. PARL achieves the highest mAP among vision-only methods at 81.9\% and the highest $AP_{50}$ of \textbf{93.7\%}. This represents a 1.2-point mAP improvement over our baseline (80.7\%) and a substantial 4.2-point gain over the powerful DINO model (77.7\%). Our vision-only PARL model significantly outperforms the multimodal LayoutLMv3 (75.4\%) by 6.5 mAP points, while using less than half the parameters (65M vs. 133M). This outcome strongly supports our central thesis regarding the efficacy of explicit structural modeling in vision-only frameworks.

\begin{table}[t]
\small
\setlength{\tabcolsep}{3pt}
\begin{tabular}{l|ccc|rr}
\toprule
\textbf{Method} & \textbf{Params} & \textbf{Latency} & \textbf{GFLOPs} & \textbf{mAP} & \textbf{AP50} \\
\midrule
\multicolumn{6}{c}{\textit{Multimodal Methods}} \\
\midrule
LayoutLMv3 & 133M & - & 64.8 & 75.4 & 92.1 \\
Cascade (m2doc) & 285M & 95.96 & 907 & \underline{86.7} & - \\
DINO (m2doc) & 256M & 87.25 & 770 & \textbf{89.0} & - \\
\midrule
\multicolumn{6}{c}{\textit{Vision-only Methods}} \\
\midrule
Faster R-CNN & 60M & 26.28 & 181 & 73.4 & - \\
Mask R-CNN & 63M & 43.08 & 290 & 73.5 & - \\
Cascade & 96M & 52.93 & 286 & 75.6 & - \\
DINO & 67M & 34.89 & 279 & 77.7 & \underline{93.5} \\
DocLayout-YOLO & 26M & 13.05 & 68 & 79.7 & 93.4 \\
DFINE-X & 62M & 22.47 & 202 & 80.7 & 93.3 \\
\hline
\textbf{PARL (ours)} & 65M & 28.65 & 217 & 81.9 & \textbf{93.7} \\
\bottomrule
\end{tabular}
\caption{Performance comparisons on DocLayNet. The best and second best results are shown in \textbf{bold} and \underline{underline}.}
\vspace{-0.7cm}
\label{tab:doclaynet_results}
\end{table}

\noindent\textbf{Fine-Grained Per-Category Analysis On D4LA.}
To gain deeper insight into the qualitative improvements offered by PARL, we performed a fine-grained evaluation on the D4LA dataset, with a detailed, per-category performance comparison presented in Table~\ref{tab:detailed_results_combined}. This analysis reveals that PARL's enhancements are not confined to a few categories but are broadly distributed across the layout taxonomy. The model achieves the best performance in \textbf{18 out of 27} categories, excelling in those characterized by well-defined structures and consistent spatial arrangements, such as \textbf{DocTitle}, \textbf{TableName}, \textbf{ListText}, \textbf{ParaText} and \textbf{Table}. This suggests that our structure-aware components (BSP-DA and GRC) are key to distinguishing visually similar but semantically different layout components. For instance, the ability to discern a \textbf{ParaTitle} from regular \textbf{ParaText} is enhanced by understanding its spatial relationship to surrounding text blocks, a capability directly fostered by our architecture.

Conversely, the analysis of lower-performing categories underscores the inherent limitations of a purely vision-based approach and clarifies the competitive landscape. The \textbf{Equation} category, for example, exemplifies a class where identification is more dependent on the semantic content of internal symbols than on visual structure. This gives a distinct advantage to multimodal competitors like VGT, which fuses both visual patterns and symbolic textual information during inference. Other challenging categories for PARL include elements like \textbf{Catalog}, which are often small, lack distinct visual features, or exhibit high variance in layout. In a different vein, another competitor, DoPTA, demonstrates impressive performance in semantically-rich categories such as \textbf{Author} and \textbf{Reference}. Its strength, however, stems from a powerful pretraining phase that embeds rich semantic cues into its visual features, a different strategy from our architectural focus.

PARL achieves the highest overall mAP by leveraging its unique strengths. In contrast to the multimodal VGT, it remains efficient and vision-only at inference. Its success stems from a novel structure-aware architecture, distinguishing it from pre-training-reliant models like DoPTA. Winning against these distinct competitors validates our core thesis: a structure-aware design is a more effective and generalizable approach to document layout analysis.

\begin{table*}
\small
\setlength{\tabcolsep}{4pt}
\small
\begin{tabular}{l c ccccccc}
\toprule
\textbf{Model} & \textbf{Modality} & \textbf{DocTitle} & \textbf{ListText} & \textbf{LetterHead} & \textbf{Question} & \textbf{RegionList} & \textbf{TableName} & \textbf{FigureName} \\
\midrule
VGT & Multimodal & 69.89 & 68.28 & \underline{83.00} & 72.53 & \underline{81.21} & 65.61 & 54.85 \\
DoPTA & Multimodal & 73.11 & \underline{72.46} & 82.07 & \textbf{77.42} & 79.32 & 67.08 & \underline{56.86} \\
Cascade (m2doc) & Multimodal & 71.47 & 69.82 & 82.51 & 72.93 & 78.95 & 65.90 & 53.12 \\
DINO (m2doc) & Multimodal & 72.15 & 70.91 & 82.78 & \underline{74.46} & 79.92 & 66.45 & 55.43 \\
DFINE-X & Vision-only & \underline{75.21} & 69.82 & \textbf{83.40} & 59.30 & 78.74 & \underline{69.40} & 51.19 \\
DocLayout-YOLO & Vision-only & 74.00 & 64.30 & 78.90 & 59.10 & 76.20 & 66.40 & 50.30 \\
\midrule
\textbf{PARL(ours)} & Vision-only & \textbf{76.08} & \textbf{74.10} & \textbf{83.40} & 69.94 & \textbf{81.45} & \textbf{71.71} & \textbf{59.20} \\
\midrule
\textbf{Model} & \textbf{Modality} & \textbf{Footer} & \textbf{Number} & \textbf{ParaTitle} & \textbf{RegionTitle} & \textbf{LetterDear} & \textbf{OtherText} & \textbf{Abstract} \\
\midrule
VGT & Multimodal & \textbf{79.00} & 82.71 & 61.11 & 64.39 & \underline{75.08} & 57.97 & 74.90 \\
DoPTA & Multimodal & 77.88 & 83.15 & 64.07 & 65.17 & 72.70 & \underline{61.25} & \textbf{78.25} \\
Cascade (m2doc) & Multimodal & 77.03 & 82.95 & 61.96 & 64.91 & 73.48 & 58.92 & 71.90 \\
DINO (m2doc) & Multimodal & 77.91 & 83.47 & 63.04 & 65.95 & 73.96 & 60.03 & 74.98 \\
DFINE-X & Vision-only & 77.47 & \underline{83.93} & 62.60 & \underline{66.50} & 73.68 & 59.79 & 74.47 \\
DocLayout-YOLO & Vision-only & 77.30 & 80.80 & \underline{65.30} & 65.20 & 67.10 & 57.90 & 71.90 \\
\midrule
\textbf{PARL(ours)} & Vision-only & \underline{78.96} & \textbf{84.74} & \textbf{66.65} & \textbf{70.55} & \textbf{76.84} & \textbf{63.00} & \underline{75.17} \\
\midrule
\textbf{Model} & \textbf{Modality} & \textbf{Table} & \textbf{Equation} & \textbf{PageHeader} & \textbf{Catalog} & \textbf{ParaText} & \textbf{Date} & \textbf{LetterSign} \\
\midrule
VGT & Multimodal & 86.40 & \textbf{49.00} & 52.28 & 49.37 & 84.89 & 67.88 & 74.01 \\
DoPTA & Multimodal & 86.90 & 32.26 & \underline{58.22} & \textbf{60.98} & 85.75 & \underline{71.40} & \textbf{76.31} \\
Cascade (m2doc) & Multimodal & 86.53 & 39.88 & 54.92 & 45.10 & 84.47 & 68.95 & 73.04 \\
DINO (m2doc) & Multimodal & 86.97 & \underline{48.15} & 56.03 & \underline{49.88} & 85.02 & 70.06 & 74.45 \\
DFINE-X & Vision-only & \underline{87.44} & 13.22 & 56.80 & 30.64 & \underline{87.05} & 68.51 & 70.62 \\
DocLayout-YOLO & Vision-only & 81.10 & 20.60 & 54.20 & 33.00 & 78.40 & 67.10 & \underline{75.50} \\
\midrule
\textbf{PARL(ours)} & Vision-only & \textbf{88.99} & 27.18 & \textbf{60.91} & 37.05 & \textbf{89.11} & \textbf{71.99} & 74.66 \\
\midrule
\textbf{Model} & \textbf{Modality} & \textbf{RegionKV} & \textbf{Author} & \textbf{Figure} & \textbf{Reference} & \textbf{PageFooter} & \textbf{PageNumber} & \textbf{mAP} \\
\midrule
VGT & Multimodal & 66.56 & 64.09 & 76.65 & \underline{84.19} & 64.14 & 58.24 & 69.19 \\
DoPTA & Multimodal & \underline{70.30} & \textbf{70.66} & 75.73 & \textbf{84.45} & 65.82 & \underline{60.64} & \underline{70.72} \\
Cascade (m2doc) & Multimodal & 67.04 & 66.95 & 76.08 & 81.97 & 65.92 & 58.95 & 69.47 \\
DINO (m2doc) & Multimodal & 68.02 & 68.03 & 77.01 & 83.05 & \underline{66.49} & 60.04 & 70.18 \\
DFINE-X & Vision-only & 66.03 & 66.96 & \underline{79.79} & 81.30 & 65.84 & 58.35 & 67.34 \\
DocLayout-YOLO & Vision-only & \textbf{72.40} & 66.10 & 75.20 & 75.90 & 60.40 & 52.70 & 69.80 \\
\midrule
\textbf{PARL(ours)} & Vision-only & 69.95 & \underline{69.05} & \textbf{81.43} & 82.64 & \textbf{68.56} & \textbf{62.00} & \textbf{70.94} \\
\bottomrule
\end{tabular}
\caption{Detailed per-class AP and overall mAP comparisons. The best and second best results are shown in \textbf{bold} and \underline{underline}.}
\label{tab:detailed_results_combined}
\vspace{-0.3cm}
\end{table*}

\begin{figure}[t]
\centering
\includegraphics[width=85mm]{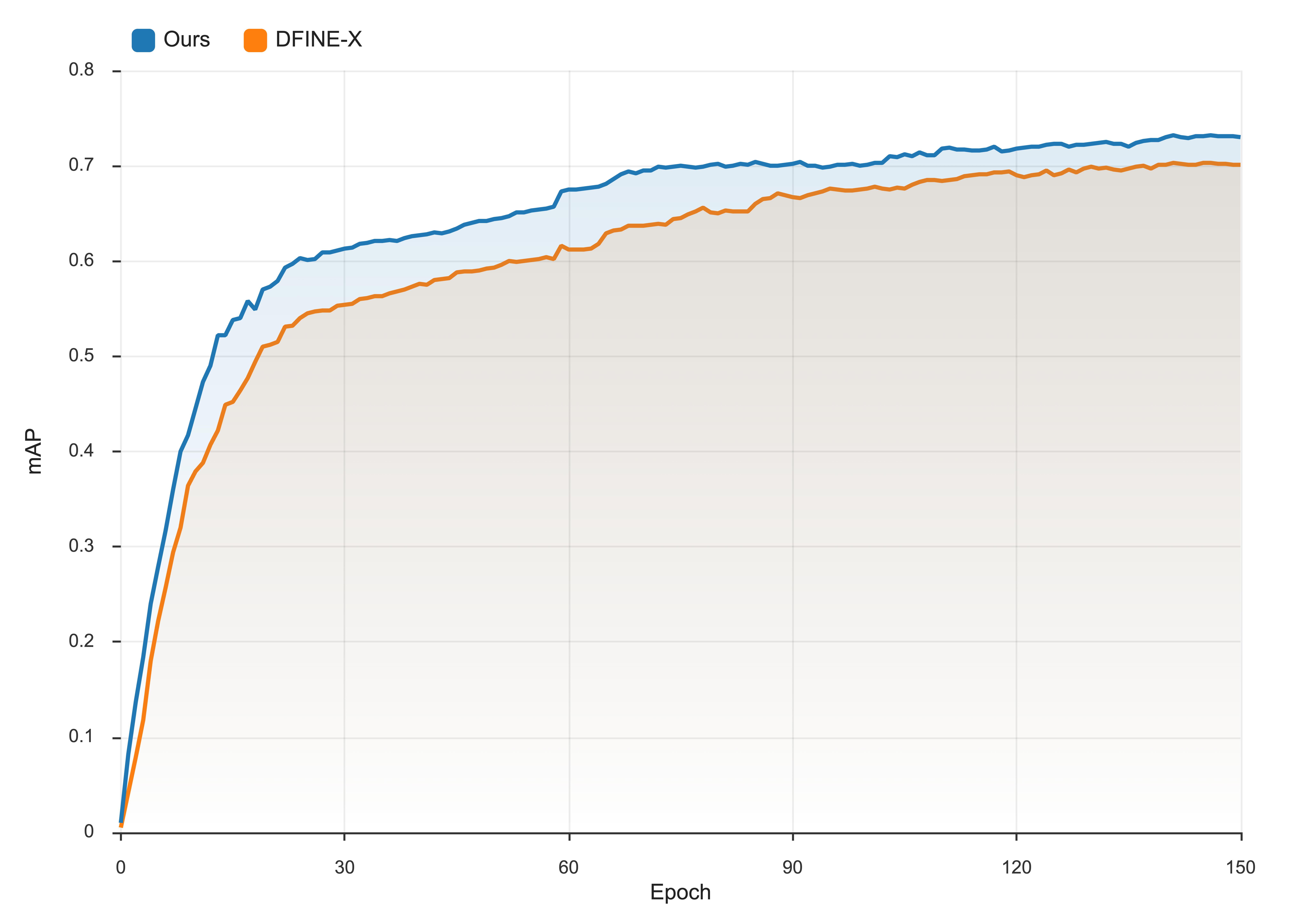} 
\caption{Convergence comparison of PARL and baseline on the M6Doc validation set.}
\label{fig:convergence}
\vspace{-0.5cm}
\end{figure}

\subsection{Ablation Study}
\label{sec:ablation_study}
To validate the effectiveness of our core components, we conducted a comprehensive ablation study on the M6Doc dataset. We systematically evaluated the individual and combined contributions of the Bidirectional Spatial Position Guided Deformable Attention (BSP-DA) module and the Graph Refinement Classifier (GRC). The results are summarized in Table~\ref{tab:ablation_study}.

Our analysis begins with the \textbf{DFINE-X model (Table~\ref{tab:ablation_study}, row 1), which serves as our baseline.} This model, which achieves an mAP of 70.3\%, employs a standard Multi-Layer Perceptron head for classification and does not include our proposed modules. It serves as the foundation for evaluating our components.

\noindent\textbf{Component Analysis.}
Effectiveness of BSP-DA. Integrating the BSP-DA module into the baseline (row 2) improves the mAP to 71.4\%, a gain of 1.1 points. This result indicates that by explicitly incorporating relative spatial relationships between object queries into the feature sampling process, BSP-DA embeds a structural prior that enhances feature representations for more precise localization.

Effectiveness of GRC. Alternatively, replacing the baseline's MLP classification head with our GRC module (row 3) boosts the mAP by 2.2 points to 72.5\%. This significant improvement underscores the importance of modeling the global layout topology. By constructing a graph and enabling predicted elements to exchange contextual information, the GRC module enhances the model's ability to disambiguate visually similar elements, thereby improving classification robustness.

\begin{table}[t]
\centering
\begin{tabular}{cc|cccc}
\toprule
\textbf{BSP-DA} & \textbf{GRC} & \textbf{AP50} & \textbf{AP75} & \textbf{Recall} & \textbf{mAP} \\
\midrule
$\times$        & $\times$        & 84.5          & 79.5          & 83.0           & 70.3          \\
$\checkmark$    & $\times$        & 85.3          & 80.4          & 83.4           & 71.4          \\
$\times$        & $\checkmark$    & 86.2          & 81.4          & 83.6           & 72.5          \\
$\checkmark$    & $\checkmark$    & \textbf{86.8} & \textbf{82.0} & \textbf{83.7}  & \textbf{72.9} \\
\bottomrule
\end{tabular}
\caption{Ablation study of PARL on the M6Doc dataset.}
\label{tab:ablation_study}
\vspace{-0.5cm}
\end{table}

Synergistic Effect. The full PARL model, integrating both BSP-DA and GRC (row 4), achieves the highest mAP of \textbf{72.9\%}, a total improvement of 2.6 points over the baseline. This result confirms a significant synergy between the two modules. The BSP-DA module provides the classifier with structurally-aware features, which the GRC module then refines through explicit reasoning about the global layout context. They facilitate a comprehensive understanding of the document structure, validating our overall design.

\noindent\textbf{Convergence Analysis.}
As illustrated in Figure~\ref{fig:convergence}, PARL demonstrates a significantly faster and more stable convergence rate than the baseline. This superior training efficiency is evident as PARL achieves the performance level that the baseline requires 30 epochs to attain in only 25 epochs, signifying a 5-epoch lead. We attribute this acceleration to the BSP-DA and GRC modules. By embedding strong structural priors, they enable the model to learn the inherent visual grammar of document layouts more efficiently, resulting in a more stable optimization path.

\section{Conclusion}
In this paper, we introduce PARL, a vision-only framework that achieves competitive performance against costly, OCR-dependent multimodal methods by explicitly modeling documents' inherent relational structure through two innovations: an attention mechanism embedding relational geometric priors and a graph-based relational classifier for refinement. Experiments demonstrate that PARL establishes a vision-only state-of-the-art while remaining highly parameter-efficient. Although our method shows superior performance, we acknowledge its limitations: the pure-vision paradigm struggles with elements defined by semantics, and page-level analysis restricts the capture of broader document context. Future work will explore lightweight vision-semantic fusion and cross-page modeling to enhance robustness and generality.

{
    \small
    \bibliographystyle{ieeenat_fullname}     \bibliography{main}
 }

\clearpage
\setcounter{page}{1}
\maketitlesupplementary

\section{Detailed Model Analysis}
\label{sec:appendix_analysis}

\subsection{Analysis of BSP-DA Gating Coefficient ($\lambda_i$)}
\label{sec:appendix_lambda}

\begin{figure}[hbt!]
    \centering
    \includegraphics[width=0.5\textwidth]{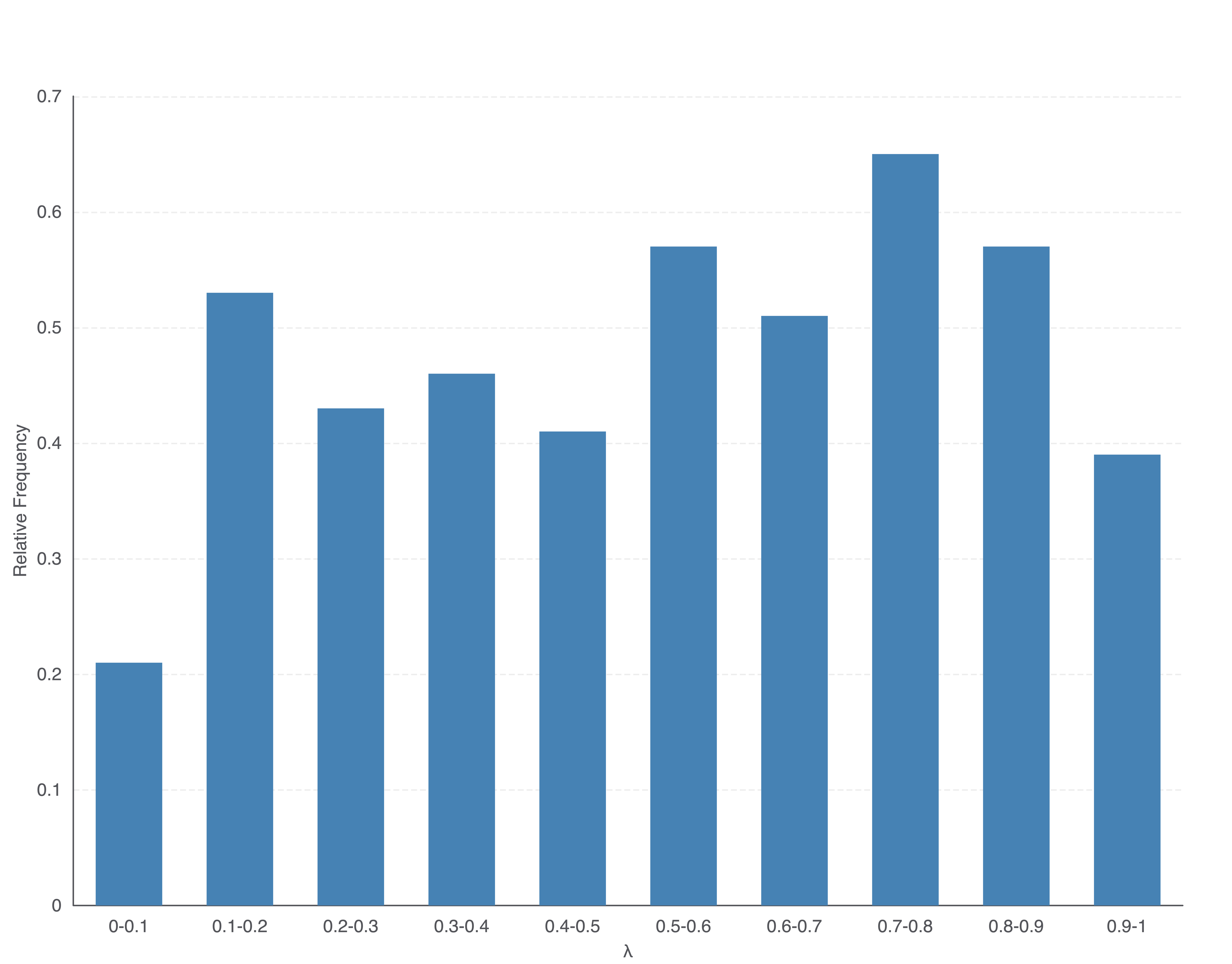}
    \caption{Distribution of the $\lambda_i$ gating coefficients on the M6Doc validation set. The wide-ranging distribution indicates that the gating mechanism is dynamically learning to balance priors and has not collapsed to 0 or 1.}
    \label{fig:lambda_histogram}
\end{figure}

The BSP-DA module introduces a learnable gating coefficient $\lambda_i$ (see Eq. 3 in the main paper) to adaptively balance the content-driven offset and the structure-driven relational adjustment. To investigate its dynamic behavior, we collected the $\lambda_i$ values predicted for all queries across the M6Doc validation set. The distribution of these values is plotted in Figure~\ref{fig:lambda_histogram}.

As shown in Figure~\ref{fig:lambda_histogram}, the $\lambda_i$ values do \textbf{not collapse} to 0 or 1. Instead, they form a \textbf{wide, dynamic distribution} across the [0, 1] range. This strongly suggests that the gating mechanism is effective and actively learning when to rely more on the content-driven base offset (when $\lambda_i \to 0$) versus the structure-driven relational adjustment (when $\lambda_i \to 1$), based on the context of each query.

\subsection{Sensitivity Analysis of GRC Neighbor Count (K)}
\label{sec:appendix_k_value}

The GRC module constructs a layout graph using a K-Nearest Neighbors (K-NN) algorithm. The choice of K is a critical hyperparameter. We conducted a sensitivity analysis on the M6Doc validation set to justify our choice, with results shown in Table~\ref{tab:k_sensitivity}.

We observe that performance peaks at $K=4$. A smaller K (e.g., K=2) may not provide sufficient contextual information. Conversely, a larger K (e.g., $K>4$) may introduce ‘‘noisy" or irrelevant neighbors into the graph, slightly degrading performance. Therefore, $K=4$ provides the optimal balance and is used in all our experiments.

\begin{table}[hbt!]
\centering
\begin{tabular}{l|ccccc}
\toprule
\textbf{K } & K=2 & \textbf{K=4 (Ours)} & K=6 & K=8 & K=10 \\
\midrule
\textbf{mAP (\%)} & 72.3 & \textbf{72.9} & 72.5 & 72.6 & 72.4 \\
\bottomrule
\end{tabular}
\caption{Sensitivity analysis for the number of neighbors (K) in the \textbf{GRC} module, evaluated on the M6Doc validation set.}
\label{tab:k_sensitivity}
\end{table}

\section{Implementation Details}
\label{sec:appendix_implementation_details}

Our model is implemented using PyTorch. Our experiments are conducted on the benchmarks established by M2Doc and M6Doc, adhering to their data and evaluation protocols to ensure a fair comparison. 

\paragraph{Benchmarking Protocol}
To ensure fair and reproducible comparisons of efficiency (e.g., Table 1 and Sec 4.2), all models were benchmarked under a strictly controlled environment using identical CPU and GPU hardware (e.g., a single NVIDIA 4090). To control for non-algorithmic bottlenecks, latency was measured on a single GPU with a batch size of 1, averaging over 50 inferences after an initial 10-iteration warmup. For baseline models, we utilized publicly available, open-source pre-trained weights where possible. For competitors where official checkpoints were not available or applicable, we re-implemented their architectures and trained them to convergence on the respective datasets to ensure an equitable performance evaluation.

To facilitate reproducibility, our source code is made publicly available on GitHub.

\section{Visual Comparison in Complex Layouts}
\label{sec:appendix_visual_comparison}

\begin{figure*}[hbt!]
    \centering
    \begin{subfigure}[b]{0.32\textwidth}
        \centering
        \includegraphics[width=\textwidth]{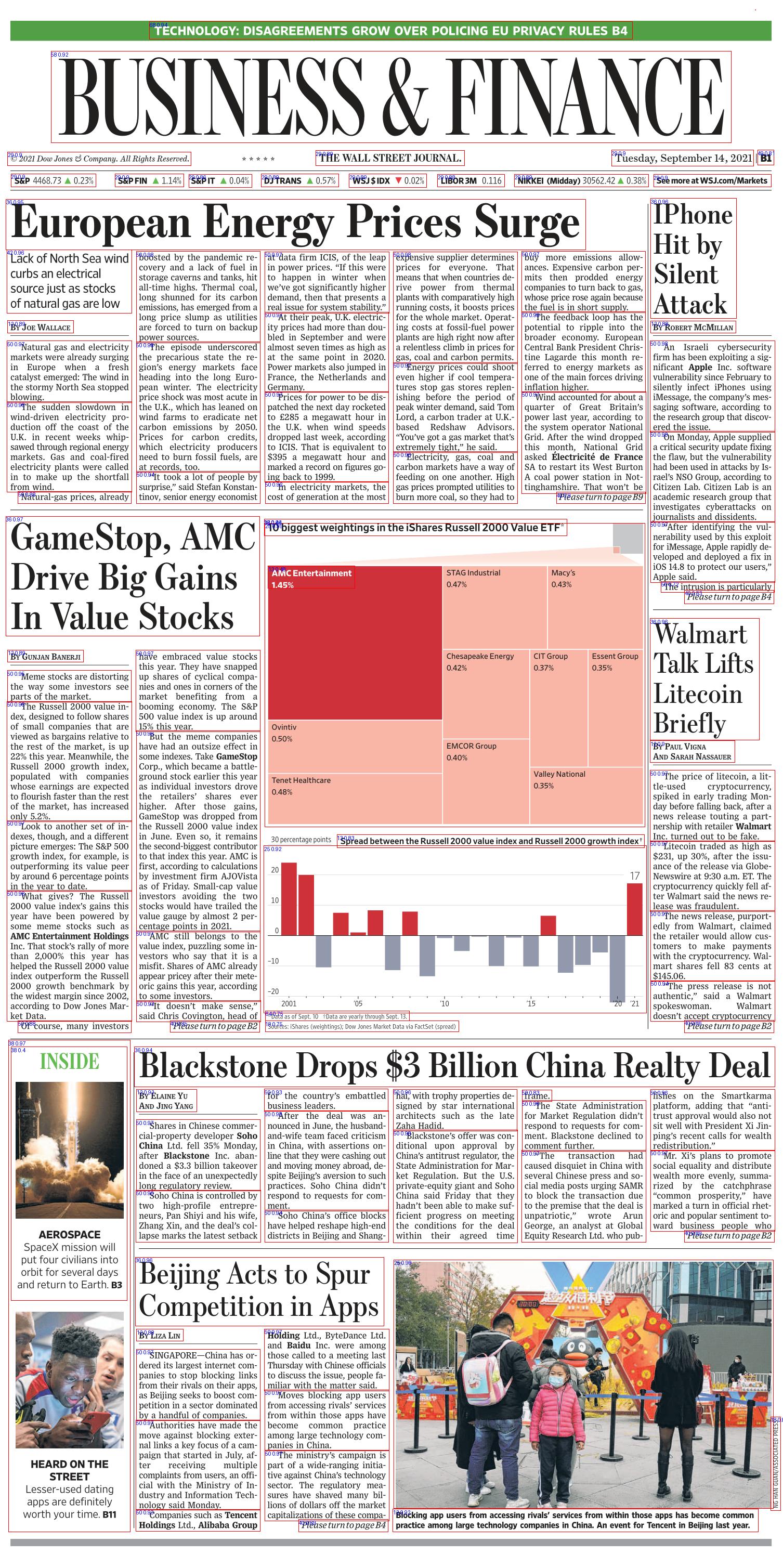}
        \caption{DINO (M2Doc)}
        \label{fig:vis_comp_a}
    \end{subfigure}
    \hfill
    \begin{subfigure}[b]{0.32\textwidth}
        \centering
        \includegraphics[width=\textwidth]{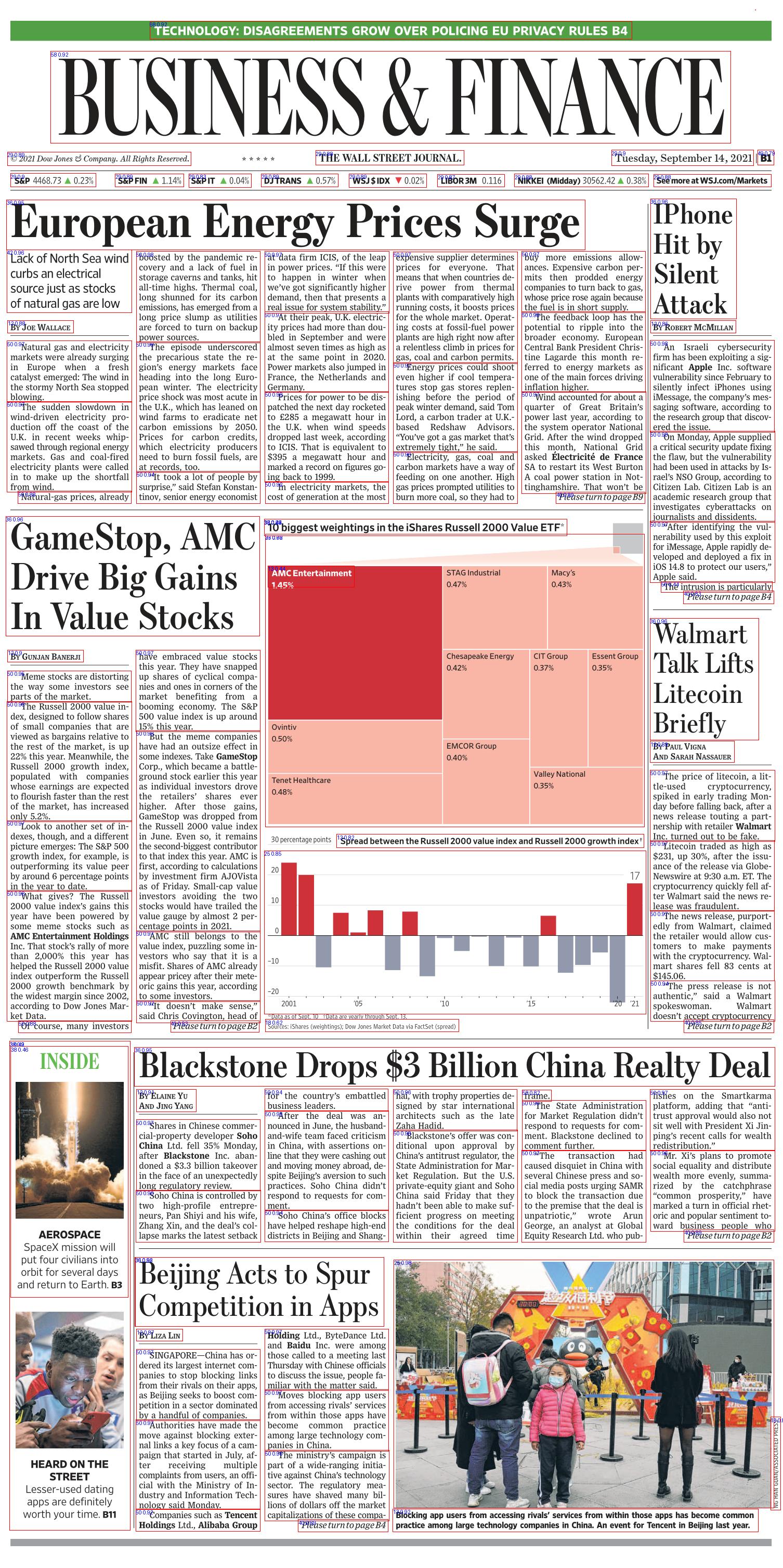}
        \caption{DFINE}
        \label{fig:vis_comp_b}
    \end{subfigure}
    \hfill
    \begin{subfigure}[b]{0.32\textwidth}
        \centering
        \includegraphics[width=\textwidth]{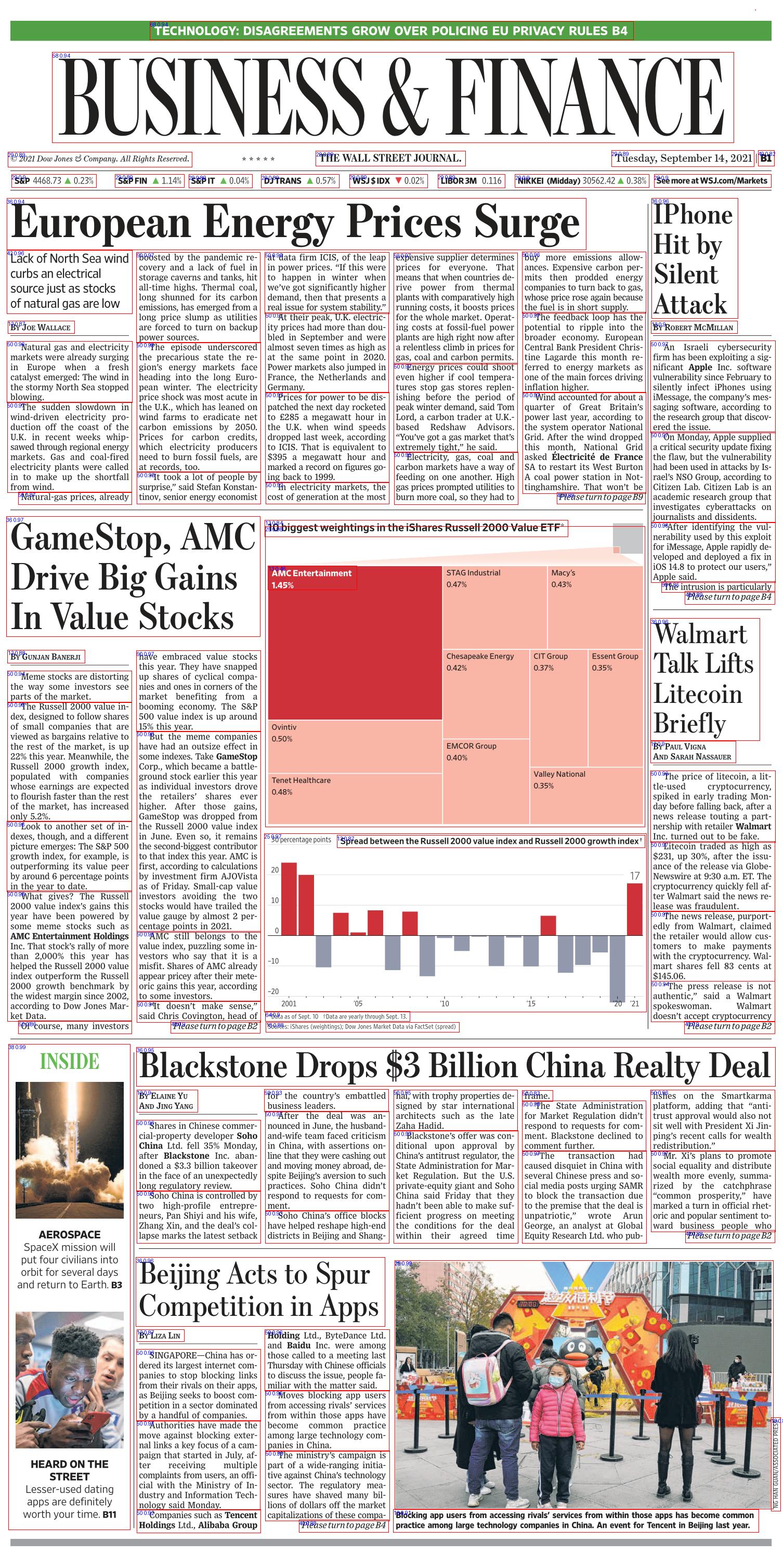}
        \caption{PARL}
        \label{fig:vis_comp_c}
    \end{subfigure}
    \caption{Qualitative comparison of different models on documents with complex layouts. From left to right: \textbf{DINO (M2Doc)}, DFINE, and PARL. Our model demonstrates superior performance in handling intricate structures.}
    \label{fig:visual_comparison}
\end{figure*}

To complement the quantitative results, Figure~\ref{fig:visual_comparison} presents a qualitative comparison that intuitively demonstrates PARL's performance advantages on challenging real-world documents. These side-by-side visualizations juxtapose the detection results of our model (PARL) against those of other leading methods, such as \textbf{DINO (M2Doc)} and DFINE. The examples specifically highlight PARL's enhanced capability to correctly identify and delineate structurally ambiguous or densely packed layout elements. This superior performance stems directly from our model's core mechanisms, namely the BSP-DA and GRC modules, which facilitate a deeper understanding of the document's inherent visual grammar via explicit positional and relational reasoning.

\section{Failure Case Analysis}
\label{sec:appendix_failure_cases}

To provide a transparent assessment of our model's limitations, we also analyzed its primary failure modes. Our qualitative review on the challenging \textbf{D4LA dataset} reveals that the majority of errors are not localization failures (i.e., failing to detect an element) but rather \textbf{inter-class confusion}.

This confusion typically occurs between layout categories that are visually similar or share ambiguous structural contexts, such as mistaking a ‘Figure' for a ‘Table', or a ‘Caption' for a ‘Paragraph'. As PARL is a vision-only model, it can be misled in scenarios where the definitive classifier is the semantic (textual) content within the element, which our model does not process. This aligns with the known limitations of non-multimodal approaches.

\begin{figure*}[hbt!]
    \centering
    \begin{subfigure}[b]{0.32\textwidth}
        \centering
        \includegraphics[width=\textwidth]{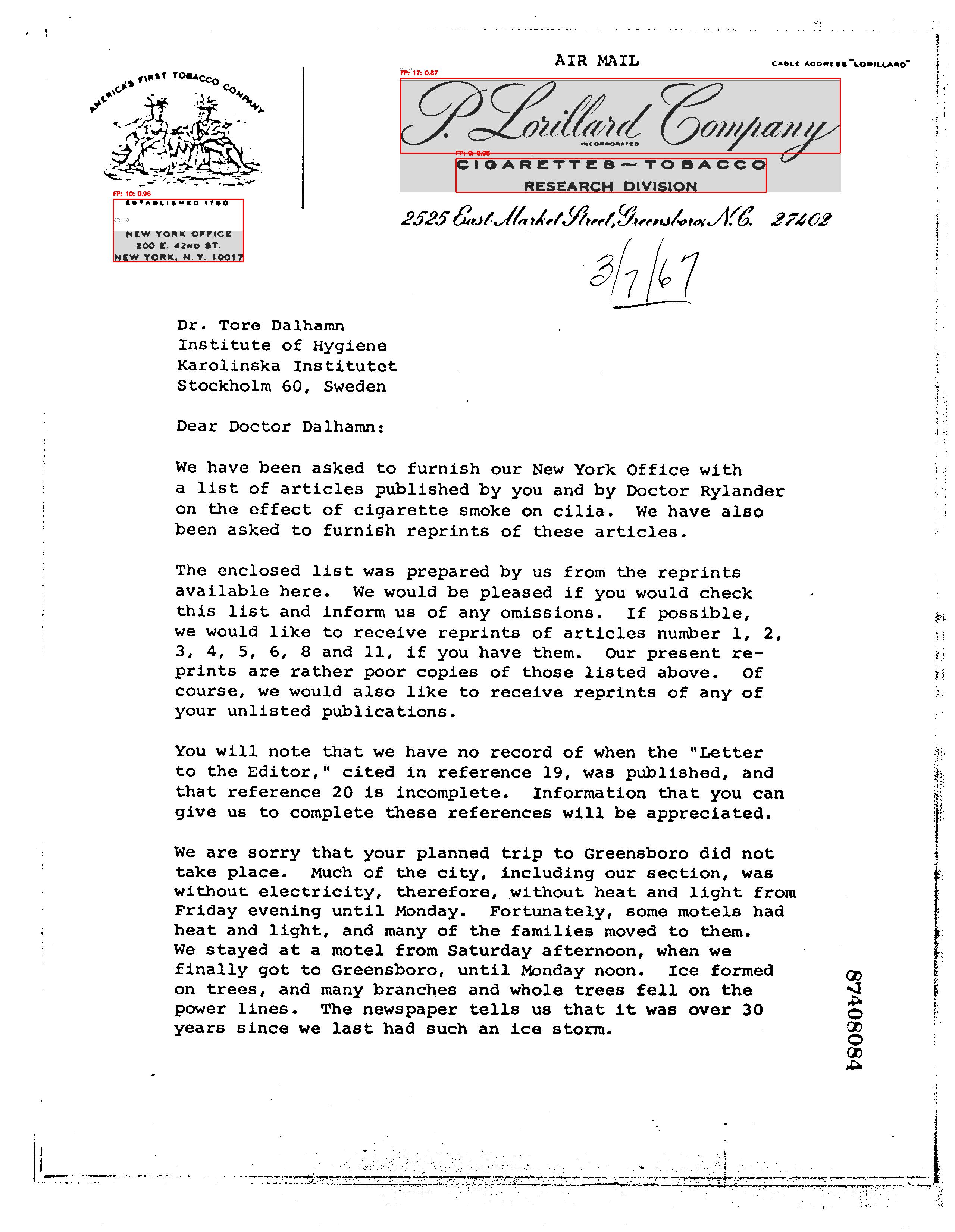}
        \caption{Example 1}
        \label{fig:fail_a}
    \end{subfigure}
    \hfill
    \begin{subfigure}[b]{0.32\textwidth}
        \centering
        \includegraphics[width=\textwidth]{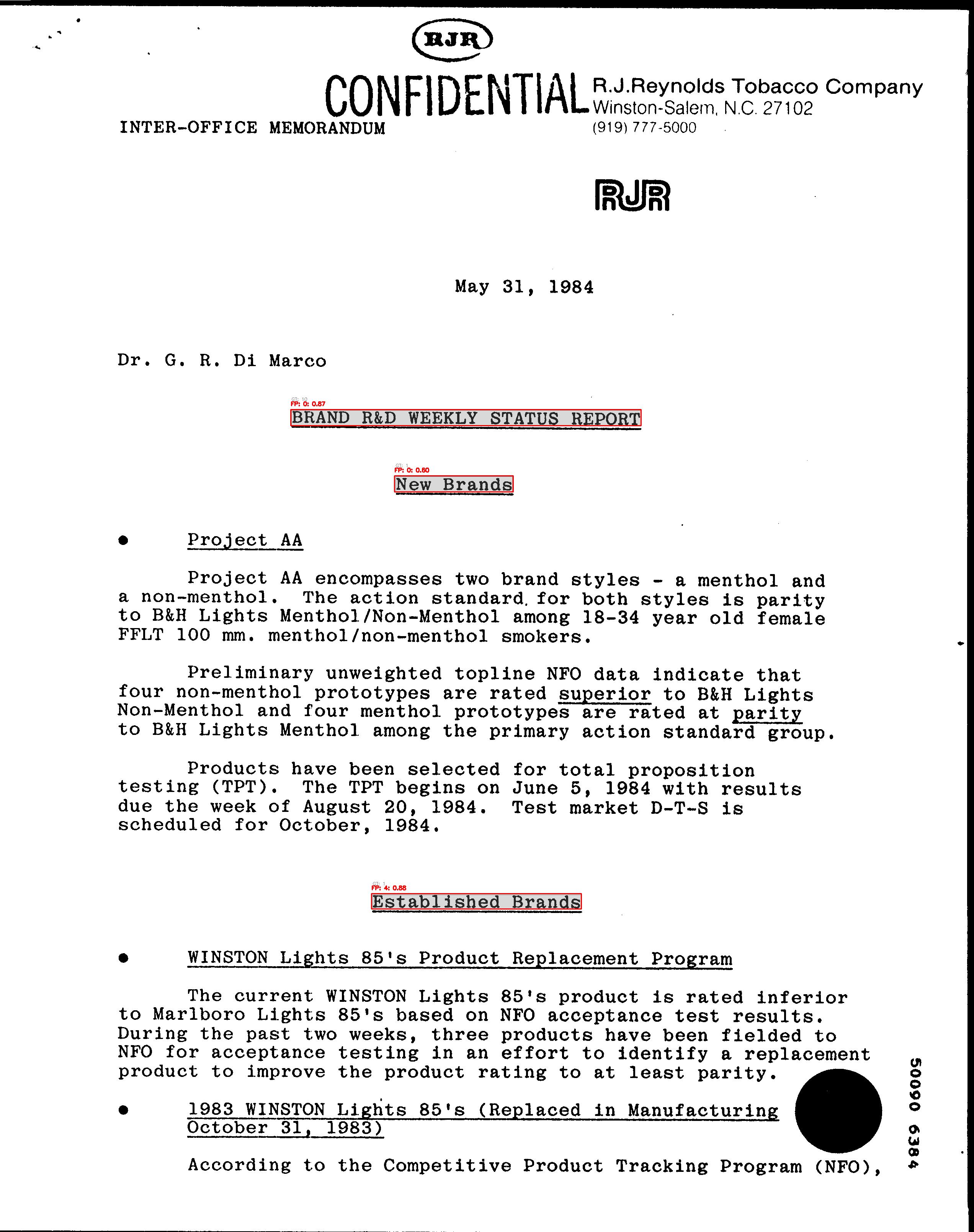}
        \caption{Example 2}
        \label{fig:fail_b}
    \end{subfigure}
    \hfill
    \begin{subfigure}[b]{0.32\textwidth}
        \centering
        \includegraphics[width=\textwidth]{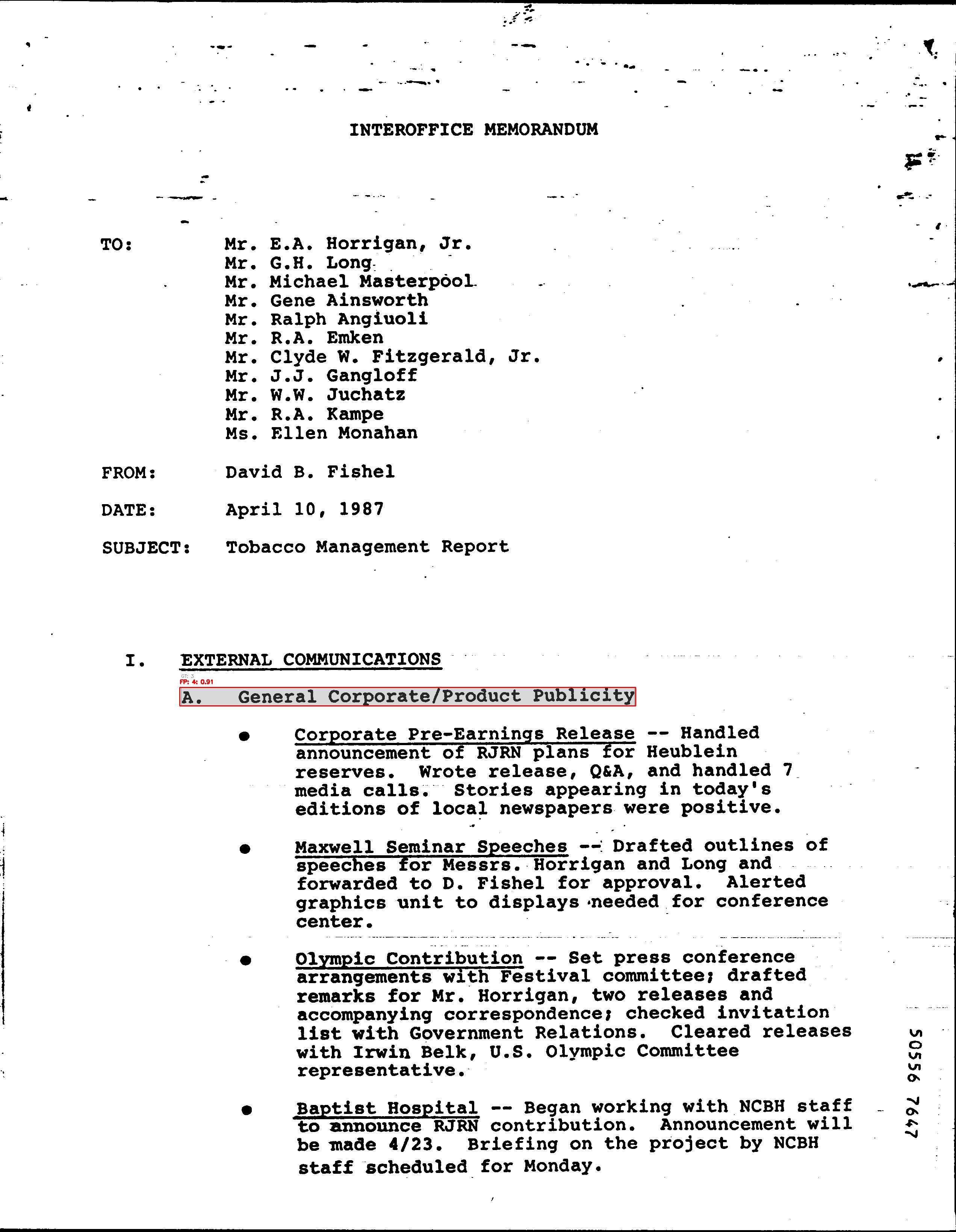}
        \caption{Example 3}
        \label{fig:fail_c}
    \end{subfigure}
    \caption{Qualitative examples of failure cases (bad cases) on the D4LA dataset. These images primarily illustrate inter-class confusion, which is the main source of error for our model on this benchmark.}
    \label{fig:failure_cases}
\end{figure*}

Figure~\ref{fig:failure_cases} presents several examples of these challenging cases from the D4LA dataset, where our model correctly localized the element but assigned an incorrect, though often structurally plausible, class label.

\end{document}